\documentclass[10pt,twocolumn,letterpaper]{article}

\usepackage{iccv}
\usepackage{times}
\usepackage{epsfig}
\usepackage{graphicx}
\usepackage{amsmath}
\usepackage{amssymb}
\usepackage{amsfonts}
\usepackage{booktabs}
\usepackage{multirow}
\usepackage{multicol}
\usepackage{mathtools,amsmath,graphicx,array}

\usepackage{graphicx}
\usepackage{float}
\usepackage{subfig}
\usepackage{subfloat}
\usepackage{caption}
\usepackage{subcaption}
\usepackage{svg}
\usepackage{bbding}
\usepackage{color}
\usepackage{graphicx}
\usepackage{amsmath}
\usepackage{amssymb}
\usepackage{booktabs}
\usepackage{makecell}

\usepackage[pagebackref=true,breaklinks=true,letterpaper=true,colorlinks,bookmarks=false]{hyperref}

\usepackage[capitalize]{cleveref}
\crefname{section}{Sec.}{Secs.}
\Crefname{section}{Section}{Sections}
\Crefname{table}{Table}{Tables}
\crefname{table}{Tab.}{Tabs.}

\iccvfinalcopy 

\def\iccvPaperID{5604} 

\ificcvfinal\fi

\begin{document}

\title{SkeletonMAE: Graph-based Masked Autoencoder for Skeleton Sequence Pre-training}

\author{Hong Yan\textsuperscript{1} \quad Yang Liu\textsuperscript{1}\thanks{Corresponding author is Yang Liu.} \quad Yushen Wei\textsuperscript{1} \quad Zhen Li\textsuperscript{2} \quad Guanbin Li\textsuperscript{1} \quad Liang Lin\textsuperscript{1}\\
\textsuperscript{1}Sun Yat-sen University \quad \textsuperscript{2}The Chinese University of Hong Kong (Shenzhen)\\
{\tt\small \{yanh36,weiysh8\}@mail2.sysu.edu.cn, 
 \{liuy856,liguanbin\}@mail.sysu.edu.cn,}\\
 {\tt\small lizhen@cuhk.edu.cn, linliang@ieee.org}
}

\maketitle
\ificcvfinal\thispagestyle{empty}\fi

\begin{abstract}
   Skeleton sequence representation learning has shown great advantages for action recognition due to its promising ability to model human joints and topology. However, the current methods usually require sufficient labeled data for training computationally expensive models, which is labor-intensive and time-consuming. Moreover, these methods ignore how to utilize the fine-grained dependencies among different skeleton joints to pre-train an efficient skeleton sequence learning model that can generalize well across different datasets. In this paper, we propose an efficient skeleton sequence learning framework, named Skeleton Sequence Learning (SSL). To comprehensively capture the human pose and obtain discriminative skeleton sequence representation, we build an asymmetric graph-based encoder-decoder pre-training architecture named SkeletonMAE, which embeds skeleton joint sequence into Graph Convolutional Network (GCN) and reconstructs the masked skeleton joints and edges based on the prior human topology knowledge. Then, the pre-trained SkeletonMAE encoder is integrated with the Spatial-Temporal Representation Learning (STRL) module to build the SSL framework. Extensive experimental results show that our SSL generalizes well across different datasets and outperforms the state-of-the-art self-supervised skeleton-based action recognition methods on FineGym, Diving48, NTU 60 and NTU 120 datasets. Additionally, we obtain comparable performance to some fully supervised methods. The code is avaliable at \url{https://github.com/HongYan1123/SkeletonMAE}
\end{abstract}

\vspace{-10pt}
\section{Introduction}
\label{sec:intro}

\begin{figure}[h]
  \centering
  \begin{center}
    \includegraphics[scale=0.37]{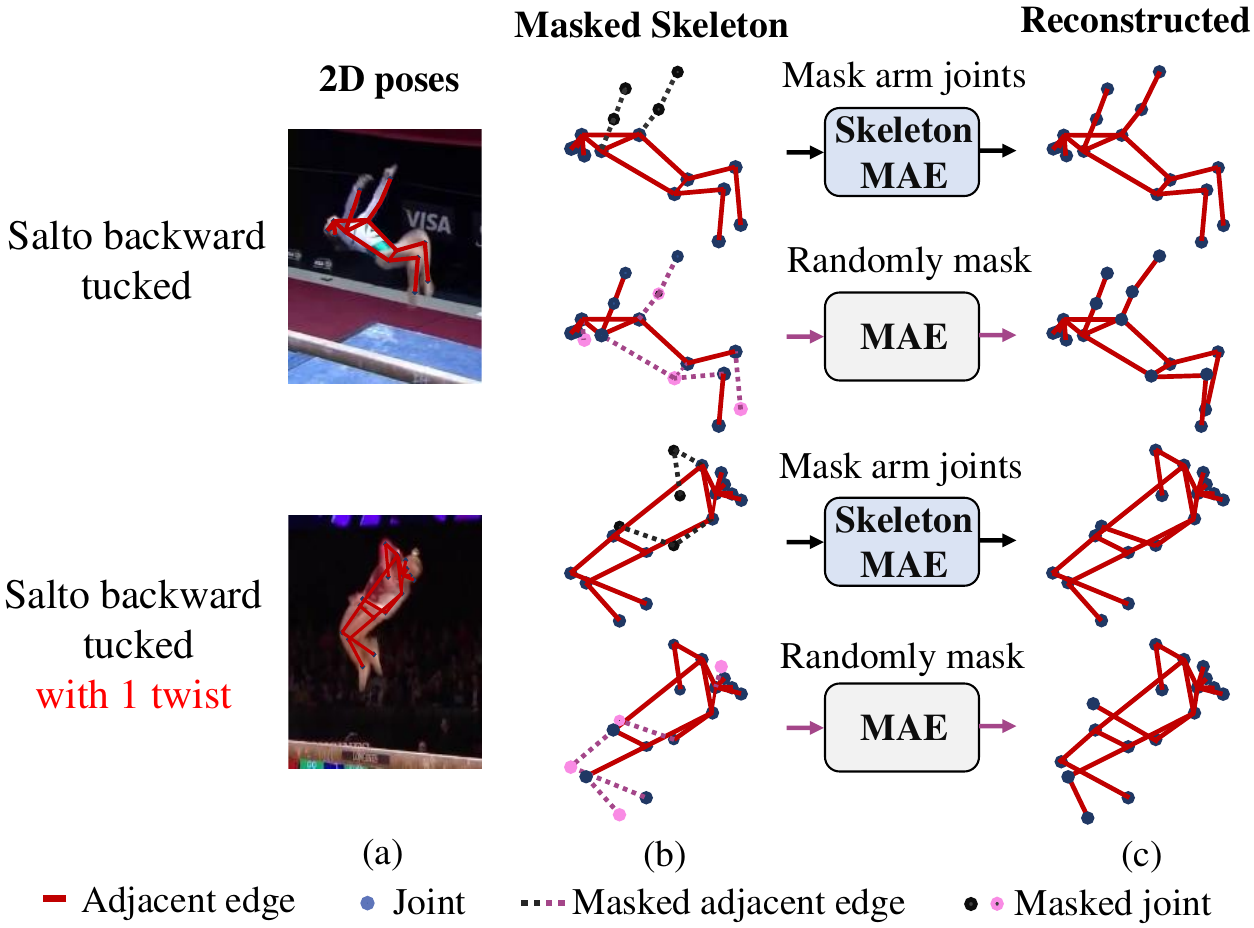}
  \end{center}
  \hfil
  \vspace{-20pt}
\caption{Traditional MAE usually uses random masking strategy to reconstruct skeleton, which tends to ignore action-sensitive skeleton regions. Differently, our proposed SkeletonMAE reconstructs the masked skeleton joints and edges based on the prior human topology knowledge, to obtain a comprehensive perception of the action.  }
\vspace{-15pt}
  \label{fig1}
\end{figure}
Human action recognition has attracted increasing attention in video understanding \cite{yan2018spatial,cho2020self,liu2020disentangling,shi2019two,liu2019deep,liu2021semantics,liu2022tcgl}, due to its wide applications \cite{carreira2017quo,feichtenhofer2019slowfast,shi2019gesture,nikam2016sign,nwakanma2021iot,yang2019gesture,lan2022audio,wang2023urban} in human-computer interaction \cite{liu2018hierarchically,CMCIR,wei2023visual}, intelligent surveillance security \cite{liu2018global,zhu2022hybrid}, virtual reality, etc. In terms of visual perception \cite{johansson1973visual,liu2022causal}, even without appearance information, humans can identify action categories by only observing the motion of joints. 
Unlike RGB videos \cite{chen2022frame, feichtenhofer2019slowfast,fan2016video,liu2018transferable}, the skeleton sequences only contain the coordinate information of the key joints of the human body, which is high-level, light-weighted, and robust against complex backgrounds and various conditions including viewpoint, scale, and movement speed \cite{duan2022revisiting,thoker2021skeleton}. Additionally, with the development of human pose estimation algorithms \cite{chu2017multi,cao2017realtime}, the localization method of human joints (i.e., key points) has made great progress and it is feasible to obtain accurate skeleton sequences. At present, the existing 2D pose estimation method is more accurate and more robust than the 3D pose estimation methods \cite{duan2022revisiting}. In \cref{fig1} (a), we visualize 2D poses estimated with HRNet \cite{sun2019deep} for two action classes on FineGym dataset \cite{shao2020finegym}. It can be seen that the 2D poses can accurately capture human skeletons and motion details.



Due to the promising ability to model multiple granularities and large variations in human motion, the skeleton sequence is more suitable to distinguish similar actions with subtle differences than the RGB data. To capture discriminative spatial-temporal motion patterns, most of the existing skeleton-based action recognition methods \cite{duan2022revisiting,yan2018spatial, chen2021channel} are fully supervised and usually require large amounts of labeled data for training elaborate models, which is time-consuming and labor-intensive. To mitigate the problem of limited labeled training data, self-supervised skeleton-based action recognition methods \cite{li20213d,guo2022contrastive,su2020predict} have attracted increasing attention recently. Some contrastive learning methods \cite{li20213d,guo2022contrastive} adopted data augmentation to generate pairs of positive and negative samples, but they rely heavily on the number of contrastive pairs. With the popularity of the encoder-decoder \cite{srivastava2015unsupervised,martinez2017human}, some methods \cite{zheng2018unsupervised,su2020predict} reconstructed the masked skeleton sequence by link reconstruction to encourage the topological closeness following the paradigm of graph encoder-decoder. However, these methods are usually good at link prediction and node clustering but are unsatisfactory in node and graph classifications. For accurate action recognition, the fine-grained dependencies among different skeleton joints (i.e., graph classifications) are essential. Therefore, previous self-supervised learning-based methods tend to ignore the fine-grained dependencies among different skeleton joints, which restricts the generalization of self-supervised skeleton representation. As shown in \cref{fig1} (b)-(c), the arm joints and edges are essential to discriminate between these two similar actions. Different from the randomly masking strategy of MAE \cite{he2022masked}, our masking strategy is action-sensitive and reconstructs specific limbs or body parts that dominate the given action class. Our SkeletonMAE utilizes prior human topology knowledge to guide the reconstruction of the masked skeleton joints and edges to achieve a comprehensive perception of the joints, topology, and action.

To address the aforementioned challenges, we propose an efficient skeleton sequence representation learning framework, named Skeleton Sequence Learning (SSL). To fully discover the fine-grained dependencies among different skeleton joints, we build a novel asymmetric graph-based encoder-decoder pre-training architecture named SkeletonMAE that embeds skeleton joint sequences in Graph Convolutional Network (GCN). The SkeletonMAE aims to reconstruct the masked human skeleton joints and edges based on prior human topology knowledge so that it can infer the underlying topology of the joints and obtain a comprehensive perception of human action. To learn discriminative spatial-temporal skeleton representation, the pre-trained SkeletonMAE encoder is integrated with the Spatial-Temporal Representation Learning (STRL) module to learn spatial-temporal dependencies. Finally, the SSL is fine-tuned on action recognition datasets. Extensive experimental results on FineGym, Diving48, NTU 60 and NTU 120 show that our SSL generalizes well across different datasets and outperforms the state-of-the-art methods significantly. Our contributions are summarized as follows: 
\begin{itemize}
\setlength{\itemsep}{0pt}
\setlength{\parsep}{0pt}
\setlength{\parskip}{0pt}
    \item  To comprehensively capture human pose and obtain discriminative skeleton sequence representation, we propose a graph-based encoder-decoder pre-training architecture named SkeletonMAE, that embeds skeleton joint sequence into GCN and utilize the prior human topology knowledge to guide the reconstruction of the underlying masked joints and topology.
    
    \item To learn comprehensive spatial-temporal dependencies for skeleton sequence, we propose an efficient skeleton sequence learning framework, named Skeleton Sequence Learning (SSL), which integrates the pre-trained SkeletonMAE encoder with the Spatial-Temporal Representation Learning (STRL) module.
    
    \item Extensive experimental results on FineGym, Diving48, NTU 60 and NTU 120 datasets show that our SSL outperforms the state-of-the-art self-supervised skeleton-based action recognition methods and achieves comparable performance compared with the state-of-the-art fully supervised methods.
\end{itemize}


\vspace{-10pt}
\section{Related Work}
\label{sec:relat}

\noindent\textbf{Action Recognition.}\quad One of the most challenging tasks for action recognition is to distinguish similar actions from subtle differences. Recently, some challenging action recognition datasets like FineGym \cite{shao2020finegym}, Diving48 \cite{li2018resound}, NTU RGB+D 60 \cite{shahroudy2016ntu} and NTU RGB+D 120 \cite{liu2019ntu} are proposed. These datasets contain a large number of challenging actions that require discriminative and fine-grained action representation learning. For example, in FineGym~\cite{shao2020finegym}, an action is divided into action units, sub-actions, or phases, and the model is required to distinguish between ``split leap with 1 turn" and ``switch leap with 1 turn". The higher inter-class similarity and a new level of granularity in the fine-grained setting make it a challenging task, which makes coarse-grained backbones and methods~\cite{feichtenhofer2019slowfast,carreira2017quo,tran2018closer,wang2021action} struggle to overcome. To tackle the more challenging fine-grained action recognition task, most of the existing works~\cite{munro2020multi,li2022dynamic} are fully supervised and consider fine-grained actions as distinct categories and supervise the model to learn action semantics. However, collecting and labeling these fine-grained actions is time-consuming and labor-intensive, which limits the generalization of a well-trained model to different datasets. To utilize unlabeled data, we propose a graph-based encoder-decoder pre-training
architecture named SkeletonMAE.

\noindent\textbf{Skeleton-based  Action Recognition.}\quad Due to the promising ability to model multiple granularities and large variations in human motion, the skeleton data is more suitable for the aforementioned action recognition task than the RGB data~\cite{chen2021learning}. Early skeleton-based action recognition methods are usually handcrafted, exploiting the geometric relationship of skeleton joints \cite{lv2006recognition,vemulapalli2014human,wang2012mining,vemulapalli2016rolling}, which greatly limits the feature representation of skeletons. Benefiting from the advantages of deep learning, some methods~\cite{zhu2016co,song2017end,song2018spatio}  utilized RNNs as the basic model, Du \etal\cite{du2015hierarchical} presented a pioneering work based on hierarchical RNNs. But RNNs easily suffer from vanishing gradients~\cite{hochreiter2001gradient}. Inspired by the booming Graph Convolutional Networks (GCN)~\cite{kipf2016semi}, Yan \etal\cite{yan2018spatial} proposed a spatial-temporal graph convolutional network to learn the spatial and temporal pattern from skeleton data. However, their manually defined topology is arduous to model the relations among joints in underlying topology. Chen \etal\cite{chen2021channel} proposed a channel-wise topology graph convolution, which models channel-wise topology with a refinement method. Duan \etal\cite{duan2022revisiting} proposed a PoseConv3D model that relies on a 3D heatmap volume instead of a graph sequence as the base representation of human skeletons. Different from previous methods that required large amounts of labeled data for training elaborate models, we utilize unlabeled skeleton sequences to pre-train a graph-based encoder-decoder named SkeletonMAE to comprehensively capture human pose and obtain discriminative skeleton sequence representation. 
 
\noindent\textbf{Self-supervised Learning for Skeleton Sequence.}\quad To learn more effective representation for unlabeled skeleton data, self-supervised learning has achieved inspiring progress recently. For contrastive learning approaches, AS-CAL~\cite{rao2021augmented} and SkeletonCLR~\cite{li20213d} applied momentum encoders for contrastive learning with single-stream skeleton sequences. AimCLR~\cite{guo2022contrastive} used an extreme data augmentation strategy to add additional hard contrastive pairs. Most contrastive learning methods adopt data augmentation to generate positive and negative pairs, but they rely heavily on the number of contrastive pairs. 
For generative learning approaches, LongT GAN~\cite{zheng2018unsupervised} proposed the encoder-decoder to reconstruct masked input sequence skeletons. Based on the LongT GAN, P\&C \cite{su2020predict} strengthened the encoder and weakened the decoder for feature representation. Wu \etal \cite{Wu2022skeletonmae} proposed a spatial-temporal masked auto-encoder framework for self-supervised 3D skeleton-based action recognition. Colorization~\cite{yang2021skeleton} used three pairs of encoder-decoder frameworks to learn spatial-temporal features from skeleton point clouds. 
Due to the limitation of the reconstruction criterion, previous generative methods usually fail to fully discover the fine-grained spatial-temporal dependencies among different skeleton joints. In our SkeletonMAE, we utilize the prior human topology knowledge to infer the skeleton sequence and obtain a comprehensive perception of the action. 

\vspace{-5pt}
\section{Methodology}
\label{sec:method}
\vspace{-5pt}

In this section, we introduce the details of Skeleton Sequence Learning (SSL), which contains two parts: 1) pre-training SkeletonMAE and 2) fine-tuning on downstream datasets based on the pre-trained SkeletonMAE.

\subsection{Pre-training SkeletonMAE} 
In this section, we introduce graph-based asymmetric encoder-decoder pre-training architecture SkeletonMAE, to learn human skeleton sequence representations without supervision. Since Graph Isomorphism Network (GIN) \cite{xu2018powerful} provides a better inductive bias, it is more suitable for learning more generalized self-supervised representation \cite{hou2022graphmae}. Therefore, we adopt GIN as the backbone of SkeletonMAE. Besides, we evaluate different backbones of SkeletonMAE in \cref{table:abla}, including GIN \cite{xu2018powerful}, GCN \cite{kipf2016semi}, and GAT \cite{velickovic2017graph}.

\vspace{-10pt}
\subsubsection{SkeletonMAE Structure}
\label{sec:Pre-tr}
Inspired by the effective representation learning by masked autoencoder (MAE) in NLP \cite{devlin2018bert}, image recognition \cite{he2022masked}, and video recognition \cite{tong2022videomae}, we focus on the human skeleton sequence and build an asymmetric graph-based encoder-decoder pre-training architecture named SkeletonMAE that embeds skeleton sequence and its prior topology knowledge in GIN. The SkeletonMAE is implemented following the paradigm of graph generative self-supervised learning. 

\begin{figure*}[h]
  \centering
  \begin{center}
    \includegraphics[width=17cm]{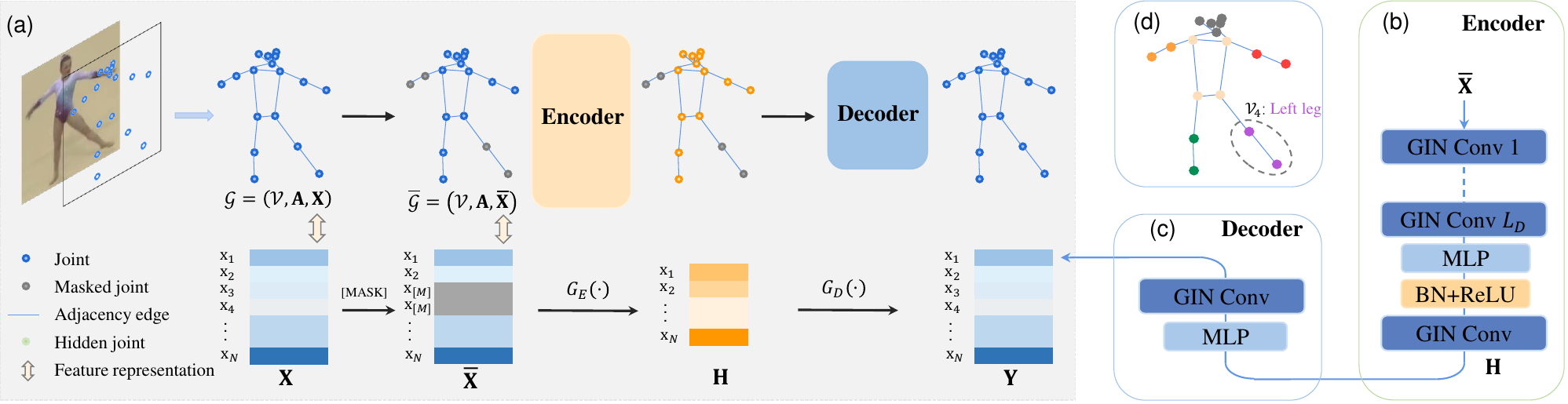}
  \end{center}
  \hfil
  
\vspace{-8pt}\caption{The details of our skeleton sequence pre-training architecture SkeletonMAE. (a) We build a GIN-based asymmetric encoder-decoder structure, to reconstruct joint features to enhance action representation ability. (b) The GIN-based encoder structure contains $L_{D}$ GIN neural network layers, to learn the joint representation spatially. (c) The decoder consists of one GIN layer, which uses the hidden features from the encoder as the input and reconstructs the original input joint features. (d) Partition the joints in the skeleton sequence according to the natural structure of the human body. 5 joints $\left\{ 
\mathcal{V}_0: \textcolor[RGB]{121,121,121}{\textbf{Head}}
\right\}$, 4 joints $\left\{\mathcal{V}_1: \textcolor[RGB]{253,222,187}{\textbf{Torso}}\right\}$, 
2 joints $\left\{\mathcal{V}_2: \textcolor[RGB]{249,64,63}{\textbf{Left arm}}, 
\mathcal{V}_3: \textcolor[RGB]{255,160,62}{\textbf{Right arm}}, 
\mathcal{V}_4: \textcolor[RGB]{183,86,215}{\textbf{Left leg}}, 
\mathcal{V}_5: \textcolor[RGB]{2,142,85}{\textbf{Right leg}}\right\}.$}
 \vspace{-5pt}
  \label{Fig2}
\end{figure*}

We follow the joint label of the Kinetics Skeleton dataset \cite{shi2019two}. Specifically, as \cref{Fig2}(d) shown, we divide all $N = 17$ joints into $R = 6$ local regions according to the natural parts of the body: $\mathcal{V}_0,...,\mathcal{V}_5$. Notably, compared to the randomly masking strategy from MAE \cite{he2022masked} to select skeleton joints, our masking strategy is action-sensitive and reconstructs specific limbs or body parts that dominate the given action class. Then, we mask these skeleton regions and make the SkeletonMAE reconstruct the masked joint features and their edges based on the adjacent joints. By reconstructing the masked skeleton joints and edges, the SkeletonMAE can infer the underlying topology of joints and obtain a comprehensive perception of the action.  


As shown in \cref{Fig2}, the SkeletonMAE is an asymmetric encoder-decoder architecture, which includes an encoder and a decoder. The encoder consists of $L_{D}$ GIN layers that map the input 2D skeleton data to hidden features. The decoder, which consists of only one GIN layer, reconstructs the hidden features under the supervision of the reconstruction criterion. According to the prior human skeleton knowledge that the human skeleton can be represented as a graph with joints as vertices and limbs as edges, we formulate the human skeleton as the following graph structure. 


The skeleton sequence of two-dimensional coordinates of $N$ human skeleton joints and $T$ frames is pre-processed in the following way. Specifically, we embed all skeleton joints and their topology into a structure $\mathcal{G}$, the skeleton structure and the joint feature are fused to obtain a joint sequence matrix $\mathbf{S}\in \mathbb{R}^{N\times T \times2}$. And then the $\mathbf{S}$ is linearly transformed to $\mathbf{S}\in \mathbb{R}^{N\times T\times D}$ with learnable parameters. We
empirically set T and D to 64. For each skeleton frame $\mathbf{X}\in \mathbb{R}^{N\times D}$ from $\mathbf{S}$, let $\mathcal{G} = (\mathcal{V},\mathbf{A},\mathbf{X})$ denote a skeleton, where $\mathcal{V} = \left\{{v}_{1},{v}_{2},......,{v}_{N}\right \}$ is the node set that contains all skeleton joints, $N=|\mathcal{V}|$ is the number of joints. The number of joints is $N = 17$. $\mathbf{A} \in \left\{0,1\right \}^{N \times N}$ is an adjacency matrix, where $\mathbf{A}_{i,j}=1$ if joints $i$ and $j$ are physically connected, otherwise 0. The feature of ${v}_{i}$ is represented as $\mathbf{x}_{i}\in \mathbb{R}^{1 \times D}$. And $\mathit{G}_{E}$, $\mathit{G}_{D}$ denote the GIN encoder and the GIN decoder, respectively. 

\vspace{-12pt}
\subsubsection{Skeleton Joints Masking and Reconstruction}
\vspace{-5pt}

 Since the prior human skeleton topology $\mathbf{A}$ is embedded (\cref{Fig2}) and we specify the aggregation of joints in \cref{sec:Pre-tr}. Inspired by the GraphMAE \cite{hou2022graphmae} that randomly reconstructs the masked graph nodes, our SkeletonMAE reconstructs the masked skeleton feature $\mathbf{X}$ based on the prior skeleton topology, rather than reconstructing graph structure $\mathbf{A}$ \cite{tang2015line,grover2016node2vec} or reconstructing both structure $\mathbf{A}$ and features $\mathbf{X}$ \cite{salehi2019graph,park2019symmetric}. 

To mask skeleton joint features, we randomly select one or more joint sets from $\mathcal{V}=\left\{\mathcal{V}_0,...,\mathcal{V}_5 \right\}$, which consists of a subset $\overline{\mathcal{V}}\subseteq  \mathcal{V}$ for masking. For the human skeleton sequence, each joint communicates with some of its adjacent joints to represent the specific action class. Therefore, it is not feasible to mask all joint sets for all action classes. 
Then, each of their features is masked with a learnable mask token vector $\left[\boldsymbol{\mathbf{MASK}}\right ]=\mathbf{x}_{\left [ \boldsymbol{\mathbf{M}}\right ]} \in \mathbb{R}^{D}$. Thus, the masked joint feature $\overline{\mathbf{x}}_{i}$ for $\mathbf{v}_{i} \in \overline{\mathcal{V}}$ in the masked feature matrix $\overline{\mathbf{X}}$ can be defined as $\overline{\mathbf{x}}_{i}=\mathbf{x}_{\left [ \boldsymbol{\mathbf{M}}\right ]}$ if $\mathbf{v}_{i} \in \overline{\mathcal{V}}$, otherwise $\overline{\mathbf{x}}_{i}= \mathbf{x}_{i}$. We set $\overline{\mathbf{X}}\in \mathbb{R}^{N\times D}$ as the input joint feature matrix of the SkeletonMAE, and each joint feature in $\overline{\mathbf{X}}$ can be defined as $\overline{\mathbf{x}}_{i}=\left \{ \mathbf{x}_{\left [ \boldsymbol{\mathbf{M}}\right ]}, \mathbf{x}_{i}\right \}$, $i= 1,2,\cdots, N$. Therefore, the masked skeleton sequence can be formulated as $\overline{\mathcal{G}} =(\mathcal{V},\mathbf{A},\overline{\mathbf{X}})$ and the objective of SkeletonMAE is to reconstruct the masked skeleton features in $\overline{\mathcal{V}}$ given the partially observed joint features $\overline{\mathbf{X}}$ with the input adjacency matrix $\mathbf{A}$. The process of SkeletonMAE reconstruction is formalized as:
\begin{equation}
  \left\{\begin{matrix}\mathbf{H}= \mathit{G}_{E}(\mathbf{A},\overline{\mathbf{X}}), \; \; \; \;  \mathbf{H}\in \mathbb{R}^{N\times D_{h}}
 \\\mathbf{Y}= \mathit{G}_{D}(\mathbf{A},\mathbf{H}),  \; \; \; \;   \mathbf{Y}\in \mathbb{R}^{N\times D}
\end{matrix}\right.,
  \label{eq:en-de}
\end{equation}
where $\mathbf{H}$ and $\mathbf{Y}$ denote the encoder output and the decoder output, respectively. The objective of SkeletonMAE can be formalized as minimizing the divergence between $\mathbf{X}$ and $\mathbf{Y}$.

\vspace{-10pt}
\subsubsection{Reconstruction Criterion}
\vspace{-5pt}
The common reconstruction criterion for masked auto-encoders is a mean squared error (MSE) in image and video tasks. However, for skeleton sequence, the multi-dimensional and continuous nature of joint features makes MSE hard to achieve promising feature reconstruction because the MSE is sensitive to dimensionality and vector norms of features \cite{friedman1997bias}. Inspired by the observation \cite{grill2020bootstrap} that the $\mathit{l}_{2}$-normalization in the cosine error maps vectors to a unit hyper-sphere and substantially improves the training stability, we utilize the cosine error as the reconstruction. 

To make the reconstruction criterion focus on harder ones among imbalanced easy-and-hard samples \cite{hou2022graphmae}, we introduce the Re-weighted Cosine Error (RCE) for SkeletonMAE. The RCE is based on the intuition that we can down-weigh easy samples’ contribution in training by scaling the cosine error with a power of $\beta\geq1$. For predictions with high confidence, their corresponding cosine errors are usually smaller than $1$ and decay faster to zero when the scaling factor $\beta>1$. Formally, given the original feature ${\mathbf{X}}\in \mathbb{R}^{N\times D}$ and the reconstructed output $\mathbf{Y}\in \mathbb{R}^{N\times D}$, the RCE is defined as:
\begin{equation}
\mathcal{L}_{\textrm{RCE}}=\sum_{\mathbf{v}_{i} \in \overline{\mathcal{V}}}^{}(\frac{1}{ |\overline{\mathcal{V}}|}-\frac{\mathbf{x}_{i}^\mathrm{T}\cdot\mathbf{z}_{i}}{\left  |\overline{\mathcal{V}}|\times \| \mathbf{x}_{i}\right \| \times  \left \| \mathbf{z}_{i}\right \|})^{\beta },
  \label{eq:STMAE}
\end{equation}
which represents the average of the similarity gap between the reconstructed feature and the input feature over all masked joints. And $\beta$ is set to 2 in our work. 


By training the SkeletonMAE to reconstruct the skeleton sequence, the pre-trained SkeletonMAE can comprehensively perceive the human skeleton structure and obtain discriminative action representation. After pre-training, the SkeletonMAE can be elegantly embedded into the Skeleton Sequence Learning (SSL) framework for fine-tuning.

\subsection{Fine-tuning for Skeleton Action Recognition}
\begin{figure}[t]
  \centering
  \begin{center}
    \includegraphics[scale=0.435]{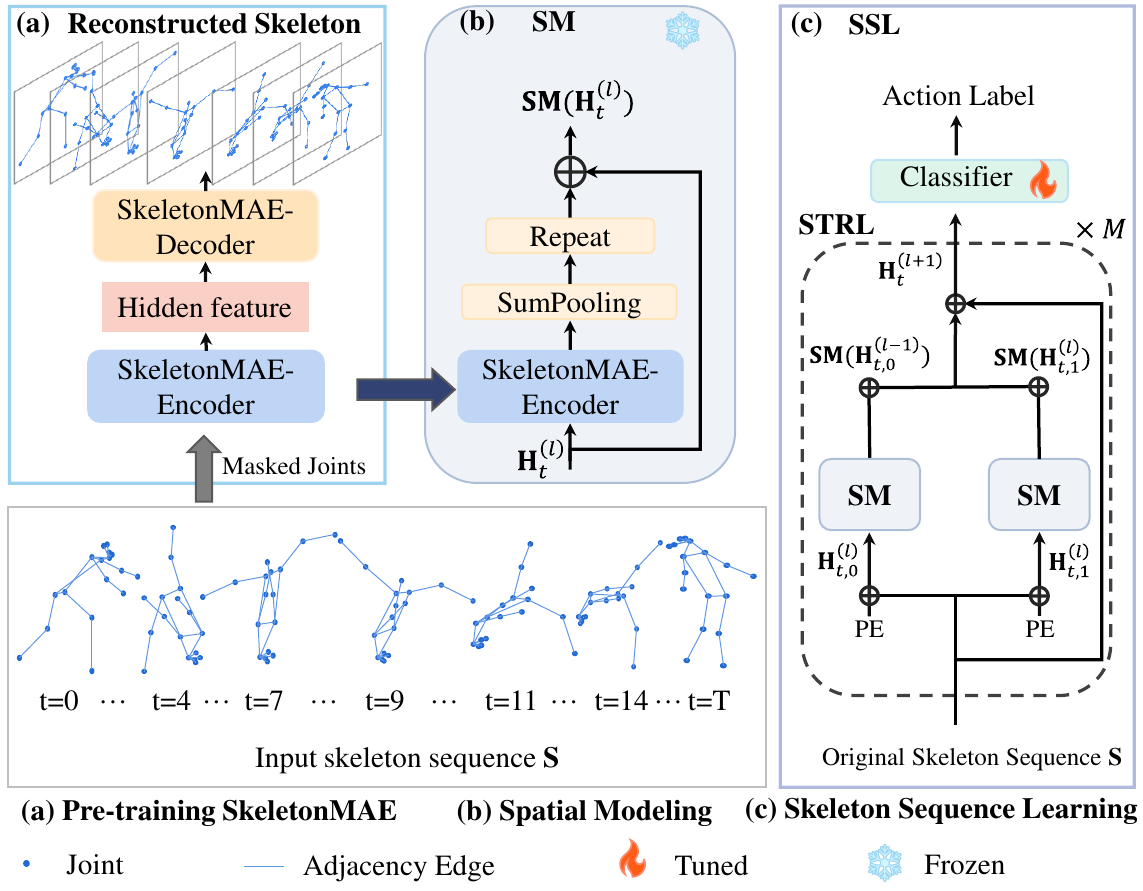}
  \end{center}
  \vspace{-8pt}
  \hfil  \vspace{-15pt}\caption{The pipeline of Skeleton Sequence Learning (SSL). (a) During pre-training, we build an encoder-decoder module named SkeletonMAE that embeds skeleton joints and its prior topology knowledge into GIN and reconstructs the underlying masked joints and topology. (b) The SM consists of the pre-trained SkeletonMAE encoder. (c)  We integrate SM structures to build the $M$-layer Spatial-Temporal Representation Learning (STRL) model and then conduct end-to-end fine-tuning. }
\vspace{-10pt}
  \label{Fig3}
\end{figure}

\vspace{-5pt}
To evaluate the SkeletonMAE's generalization ability for skeleton action recognition, we build a complete skeleton action recognition model named Skeleton Sequence Learning
(SSL), based on the pre-trained SkeletonMAE. To capture multiple-person interaction, we integrate two pre-trained SkeletonMAE encoders to build the Spatial-Temporal Representation Learning (STRL) module, as shown in \cref{Fig3}(b)-(c). The entire SSL consists of an $M$-layer STRL model and a classifier. The SSL model is finally fine-tuned on skeleton action recognition datasets with cross-entropy loss to recognize actions.

\vspace{-10pt}
\subsubsection{Spatial-Temporal Representation Learning}
\vspace{-5pt}
\label{sec:st}
The STRL contains two pre-trained SkeletonMAE encoders for Spatial Modeling (SM). The input of SM is skeleton sequence $\mathbf{S}$. The output of SM is connected with the input by $1\times1$ convolution for residual connection (\cref{Fig3} (b)). 

As shown in \cref{Fig3} (c), the input skeleton sequence $\mathbf{S}\in \mathbb{R}^{N\times T \times D}$ is firstly added with the learnable temporal position embedding $\textrm{PE}$ to obtain the skeleton sequence feature $\mathbf{H}_{t}^{(\mathit{l})}\in \mathbb{R}^{P\times N\times {D}^{(\mathit{l})}}$. To model multiple human skeleton interactions, we obtain two individual features ($P=2$) for two persons  $\mathbf{H}_{t,0}^{(\mathit{l})}\in \mathbb{R}^{N \times {D}^{(\mathit{l})}}$ and  $\mathbf{H}_{t,1}^{(\mathit{l})}\in    \mathbb{R}^{N \times {D}^{(\mathit{l})}}$ from $\mathbf{H}_{t}^{(\mathit{l})}$. Here, we take the joint feature of the 0-\textrm{th} person as an example, the operation of the 1-\textrm{th} person is implemented similarly. We send the joint representation $\mathbf{H}_{t,0}^{(\mathit{l})}$ and prior knowledge of the joint $\widetilde{\mathbf{A}}$ into the SM module, 

\begin{equation}
\begin{split}
\textrm{SM}(\mathbf{H}_{t,0}^{(\mathit{l})})=\textrm{Repeat}(\textrm{SP}(\mathit{G}_{E}\left (\widetilde{\mathbf{A}}, \mathbf{H}_{t,0}^{(\mathit{l})}
 \right));N)\oplus  
 \mathbf{H}_{t,0}^{(\mathit{l})},
  \label{eq:ST_embeding}
  \end{split}
\end{equation}
where ${G}_{E}$ is the SkeletonMAE encoder, $\textrm{SP}(\cdot{})$ denotes the sum-pooling,  $\textrm{Repeat} (\cdot{};N)$ means repeating the single joint into $N$ joints representations after sum-pooling and then connect it with the $\mathbf{H}_{t,0}^{(\mathit{l})}$ residual to get the global joint representation $\textrm{SM}(\mathbf{H}_{t,0}^{(\mathit{l})})$. In this way, the SM module can obtain global information through a single joint representation, and constrain some joint features through all joint representations. Similarly, $\textrm{SM}(\mathbf{H}_{t,1}^{(\mathit{l})})$ is obtained in the same way. As shown in \cref{Fig3}(c), we get the joint features $\textrm{SM}(\mathbf{H}_{t}^{(\mathit{l})})$ that contains the action interaction bewtween the 0-\textrm{th} person  and the 1-\textrm{th} person. According to the update rules of graph convolution \cite{kipf2016semi}, we can obtain  $\mathbf{H}_{t}^{(\mathit{l}+1)}$ from $\mathbf{H}_{t}^{(\mathit{l})}$ in a multi-layer GCN. For more details, please refer to the Supplementary in Section D. The final skeleton sequence representation is defined as follows: 
\vspace{-5pt}
\begin{equation}
\mathbf{H}_{t}^{(\mathit{l}+1)}=\sigma\left ( \textrm{SM}(\mathbf{H}_{t}^{(\mathit{l})}) \mathbf{W}^{(\mathit{l})}\right ).
  \label{eq:GC_updata_new}
  \vspace{-5pt}
\end{equation}
where $\mathbf{W}^{(\mathit{l})}$ denotes the trainable weight matrix in the $l^{th}$ layer, $\sigma(\cdot)$ denotes the ReLU activation function. 
Then, we adopt the widely-used multi-scale temporal pooling \cite{chen2021channel,liu2017enhanced} to get the final output. Finally, a classifier consisting of MLP and softmax predicts the action class.

\section{Experiments} 
\label{sec:experiments}
All experiments are conducted with a single modality (2D pose) and evaluated on the corresponding train/test sets.

\subsection{Datasets}
We evaluate our SSL on four benchmark datasets FineGym \cite{shao2020finegym}, Diving48 \cite{li2018resound}, NTU RGB+D 60 \cite{shahroudy2016ntu} and NTU RGB+D 120 \cite{liu2019ntu} in the mainstream skeleton action recognition task. For all datasets except FineGym, we follow the pre-processing protocol provided by \cite{duan2022revisiting} to obtain the skeleton sequence from the 2D pose estimator. The pre-processing adopts a Top-Down approach for pose extraction, where the detector is Faster-RCNN \cite{ren2015faster} with ResNet50 backbone and the pose estimator is HRNet \cite{sun2019deep} pre-trained on COCO-keypoint \cite{lin2014microsoft}. To make a fair comparison, we added pixel noise to the joint during training, making the original joint confidence rate unreliable, thus we do not use the originally fixed threshold.

\begin{table}[t]
\renewcommand\arraystretch{1.0}
\small
\centering
\setlength{\tabcolsep}{0.2mm}{
\begin{tabular}{lc|c}
\hline  
\multicolumn{1}{l|}{ Method} &
 Modality&  Mean Acc. (\%)  \\ \hline
\multicolumn{1}{l|}{\textbf{Fully Supervised}}                 &                        &                                                                           \\
\multicolumn{1}{l|}{I3D \cite{carreira2017quo}}                   & RGB                & 64.4                                                                           \\ 
\multicolumn{1}{l|}{ST-S3D \cite{xie2018rethinking}}                   & RGB                & 72.9                                                                           \\ 
\multicolumn{1}{l|}{TSN \cite{wang2016temporal}}                    & RGB+Flow                & 79.8                                                                           \\
\multicolumn{1}{l|}{TRNms \cite{zhou2018temporal}}                    & RGB+Flow                & 80.2                                                                           \\ 
\multicolumn{1}{l|}{TSM \cite{lin2019tsm}}                   & RGB+Flow                & 81.2                                                                           \\ 
\multicolumn{1}{l|}{GST-50 \cite{luo2019grouped}}                 & RGB                       & 84.6                                                                           \\ 
\multicolumn{1}{l|}{MTN \cite{leong2021joint}}                 & RGB                       & 88.5                                                                           \\ 
\multicolumn{1}{l|}{LT-S3D \cite{xie2018rethinking}}                & RGB                       & 88.9                                                                           \\ 
\multicolumn{1}{l|}{TQN \cite{zhang2021temporal}}                    & RGB+Text                  & 90.6                                                                           \\ 
\multicolumn{1}{l|}{PYSKL \cite{duan2022revisiting}}                  & Skeleton                  & \text{93.2}                                                                  \\ 
\multicolumn{1}{l|}{PYSKL \cite{duan2022revisiting}}                 & RGB+Skeleton+Limb                   & \textbf{95.6}                                                                  \\ \hline
\multicolumn{1}{l|}{\textbf{Unsupervised Pre-train}}                 &                        &                                                                           \\
\multicolumn{1}{l|}{SaL  \cite{misra2016shuffle}}                 & RGB                       & 42.7                                                                          \\
\multicolumn{1}{l|}{TCC  \cite{dwibedi2019temporal}}                 & RGB                       & 45.6                                                                          \\
\multicolumn{1}{l|}{GTA \cite{hadji2021representation}}                 & RGB                       & 49.5                                                                           \\
\multicolumn{1}{l|}{CARL \cite{chen2022frame}}                 & RGB                       & 60.4                                                                          \\

\hline
\multicolumn{1}{l|}{\textbf{SSL (ours)}}                 & Skeleton                  & \textbf{91.8}           
            \\ \hline
\end{tabular}}
 \vspace{-5pt}
 \caption{ The comparison with the state-of-the-art unsupervised pre-train and supervised methods on FineGym.}
  \label{tab: finetue_gym}
\end{table}

\noindent\textbf{FineGym}  is a large-scale fine-grained action recognition dataset for gymnastic videos, which contains 29K videos of 99 fine-grained gymnastics action classes, which requires action recognition methods to distinguish different sub-actions in the same video. In particular, it provides temporal annotations at both action and sub-action levels with a three-level semantic hierarchy. We follow the method \cite{duan2022revisiting} to extract the skeleton data from the 2D pose estimator.

\noindent\textbf{Diving48} is a challenging fine-grained dataset that focuses on complex and competitive sports content analysis. It is formed of over 18k video clips from competitive dive championships that are distributed over 48 fine-grained classes with minimal biases. The difficulties of the dataset lie in that actions are similar and differ in body parts and their combinations which require the model to capture details and motion in body parts and combine them to perform classification. We report the accuracy on the official train/test split.

\noindent\textbf{NTU RGB+D 60 and 120.}  \quad NTU RGB+D is a large-scale skeleton-based action recognition dataset, where NTU 60 contains 56,880 skeleton sequences and 60 action classes. NTU 120 has 114,480 skeleton sequences and 120 action categories. The NTU 60 and 120 datasets have a large amount of variation in subjects, views, and backgrounds. 
\subsection{Implementation Details} 

In this paper, our SkeletonMAE is optimized by the Adaptive Moment Estimation (Adam) with the initial learning rate as 1.5$e^{-4}$ and the PReLU is the activation function. The batch size is 1024 and the training epoch is 50. At the fine-tuning stage, we use the Stochastic Gradient Descent (SGD) with momentum (0.9) and adopt the warm-up strategy for the first 5 epochs. The total fine-tuning epochs are 110. The learning rate is initialized to 0.1 and is divided by 10 at the 90 epoch and the 100 epoch. And we employ 0.1 for label smoothing. We use a large batch size of 128 to facilitate training our attention mechanism and enhancing the model's perception for all human action classes. Both our pre-training and fine-tuning models are implemented by PyTorch \cite{paszke2017automatic}, and our SSL is trained on a single NVIDIA GeForce RTX 2080Ti GPU. For more details of implementation, please refer to the Supplementary in Section A.

\noindent\textbf{Pre-training and Fine-tuning Setting.} \quad For each dataset, the SkeletonMAE encoder is pre-trained with unlabeled data from the training set. Then, we load the learned parameter weights to fine-tune the SSL model.

\noindent\textbf{Evaluation Metrics.} \quad To make a fair comparison, we follow previous methods \cite{duan2022revisiting,chen2022frame,guo2022contrastive} and report the Mean Top-1 accuracy(\%) on FineGym dataset and Top-1 accuracy(\%) on Diving48, NTU 60, and NTU 120 datasets. 

\begin{table}[]
\renewcommand\arraystretch{1.0}
\small
\centering
\setlength{\tabcolsep}{1.5mm}{
\begin{tabular}{lccc}
\hline
Method                                                                      & Pre-train   & GFLOPs            & Acc.(\%)        \\ \hline
\multicolumn{1}{l|}{\textbf{Fully Supervised}}                                    & \multicolumn{1}{l}{} & \multicolumn{1}{l|}{}      & \multicolumn{1}{l}{} \\
\multicolumn{1}{l|}{TSN \cite{li2018resound}}                  & ImageNet             & \multicolumn{1}{c|}{-}     & 16.8                 \\
\multicolumn{1}{l|}{TRN \cite{kanojia2019attentive}}           & ImageNet             & \multicolumn{1}{c|}{-}     & 22.8                 \\
\multicolumn{1}{l|}{P3D \cite{luo2019grouped}}                 & ImageNet             & \multicolumn{1}{c|}{-}     & 32.4                 \\
\multicolumn{1}{l|}{C3D \cite{luo2019grouped}}                 & ImageNet             & \multicolumn{1}{c|}{-}     & 34.5                 \\
\multicolumn{1}{l|}{CorrNet \cite{luo2019grouped}}                  & -                    & \multicolumn{1}{c|}{74.8}  & 35.5                 \\
\multicolumn{1}{l|}{CorrNet-R101 \cite{wang2020video}}              & ImageNet             & \multicolumn{1}{c|}{187.3} & 38.2                 \\
\multicolumn{1}{l|}{MG-TEA-ResNet50 \cite{zhi2021mgsampler}}        & ImageNet             & \multicolumn{1}{c|}{-}     & 39.5                 \\
\multicolumn{1}{l|}{GSM \cite{sudhakaran2020gate}}                  & ImageNet             & \multicolumn{1}{c|}{107.4} & 40.3                 \\
\multicolumn{1}{l|}{TSM-R50 \cite{kwon2021learning}}                & ImageNet             & \multicolumn{1}{c|}{153.8} & 41.6                 \\
\multicolumn{1}{l|}{TMF \cite{wang2021tmf}}                         & ImageNet             & \multicolumn{1}{c|}{-}     & 42.2                 \\ \hline
\multicolumn{1}{l|}{\textbf{Unsupervised Pre-train}}                               &                      & \multicolumn{1}{c|}{}      &                      \\
\multicolumn{1}{l|}{RESOUND-C3D \cite{li2018resound}}               & K400                 & \multicolumn{1}{c|}{-}     & 16.4                 \\
\multicolumn{1}{l|}{Jenni \etal\cite{jenni2021time}} & K400                 & \multicolumn{1}{c|}{-}     & 29.9                 \\
\multicolumn{1}{l|}{\textbf{SSL (ours)}}                                                   & Diving48             & \multicolumn{1}{c|}{42.8}  & \multicolumn{1}{c}{\textbf{34.1}} \\ \hline
\end{tabular}}
 \vspace{-5pt}
 \caption{ The comparison with the unsupervised pre-train and supervised methods on the Diving48 dataset.}
  \label{tab: diving48}
\end{table}





\vspace{-3pt}
\subsection{Downstream Evaluation} 
For a fair comparison, we compare our SSL with other models without pre-training on additional large-scale action datasets, \textit{e.g.}, Kinetics \cite{kay2017kinetics} or Sports1M \cite{karpathy2014large}.
The comparison results on FineGym, Diving48, and NTU 60 \& 120 datasets are shown in \cref{tab: finetue_gym}, \cref{tab: diving48}, and \cref{tab: finetune_ntu}, respectively. 

\begin{table*}[h]\renewcommand\arraystretch{1.0}
\small
\centering
\setlength{\tabcolsep}{0.88mm}{
\begin{tabular}{l|cccc|cc|cc}
\hline
\multirow{2}{*}{Method}  & \multirow{2}{*}{Backbone} & \multirow{2}{*}{Supervision} & \multirow{2}{*}{Joint Number}
& \multirow{2}{*}{2D Skeleton} & \multicolumn{2}{c|}{NTU 60 } & \multicolumn{2}{c}{NTU 120} \\ \cline{6-9} & & & &                  & X-sub (\%)          & X-view (\%)          & X-sub (\%)           & X-set (\%)           \\ \hline
ST-GCN \cite{yan2018spatial} & GCN   & Fully Supervised   &  25    & \XSolidBrush     &  81.5    & 88.3         & -
   & -        \\ 
AS-GCN \cite{li2019actional} &  GCN  & Fully Supervised   &  25    & \XSolidBrush     &  86.8    & 94.2         & -  
  & -       \\
2s-AGCN \cite{shi2019two} & GCN   & Fully Supervised   &  25    & \XSolidBrush     &  88.5    & 95.1         &  82.9
   &   84.9     \\
Shift-GCN \cite{cheng2020skeleton} & GCN   & Fully Supervised   &  25    & \XSolidBrush     &  90.7    & 96.5         & 85.9 
   &  87.6      \\
MS-G3D \cite{liu2020disentangling} & GCN   & Fully Supervised   &  25    & \XSolidBrush     & 91.5    & 96.2         &  86.9
   &  88.4      \\
CTR-GCN \cite{chen2021channel} & GCN   & Fully Supervised   &  25    & \XSolidBrush     & 92.4    & \textbf{96.8}         &  \textbf{88.9}
   &  \textbf{90.6}      \\
PYSKL \cite{duan2022revisiting} & CNN   & Fully Supervised   &  17    & \Checkmark     & \textbf{93.7}    & 96.6         &  86.0
   &  89.6      \\
\hline
SkeletonCLR \cite{li20213d}  &  ST-GCN      & Unsupervised Pre-train      &   25      & \XSolidBrush                            & 82.2           & 88.9           & 73.6           & 75.3           \\ 
CrosSCLR   \cite{li20213d} &  ST-GCN        & Unsupervised Pre-train        &   25        &  \XSolidBrush                         & 86.2           & 92.5           & 80.5          & 80.4           \\ 
Wu \etal \cite{Wu2022skeletonmae} & STTFormer         & Unsupervised Pre-train   &  25      &                                   \XSolidBrush   & 86.6           & 92.9           & 76.8           & 79.1           \\ 
AimCLR  \cite{guo2022contrastive} & ST-GCN         & Unsupervised Pre-train   &  25      &                                   \XSolidBrush   & 86.9           & 92.8           & 80.1           & 80.9          \\ 
3s-PSTL \cite{zhou2023self} & ST-GCN  & Unsupervised Pre-train  &  25 & \XSolidBrush  & 87.1   & 93.9  & \underline{81.3}  & \underline{82.6}    \\
Colorization \cite{yang2021skeleton}  & DGCNN        &  Unsupervised Pre-train   & 25         & \XSolidBrush                                     & \underline{88.0}           & \underline{94.9}           & -           & -          \\
\hline
\textbf{SSL(ours)}&   STRL       &   Unsupervised Pre-train        & 17    & \Checkmark                                     & \textbf{92.8}(\textcolor[RGB]{147,72,51}{$\uparrow$ 4.8})  & \textbf{96.5}(\textcolor[RGB]{147,72,51}{$\uparrow$ 1.6})  & \textbf{84.8}(\textcolor[RGB]{147,72,51}{$\uparrow$ 3.5})  & \textbf{85.7}(\textcolor[RGB]{147,72,51}{$\uparrow$ 3.1})  \\ \hline
\end{tabular}}
 \caption{The comparison with state-of-the-art unsupervised pre-train and supervised methods on NTU 60 and NTU 120 datasets. `\underline{~~~}' means the method with the second-best performance under unsupervised pre-training. }
    \label{tab: finetune_ntu}
\end{table*}

\noindent\textbf{Results on FineGym Dataset.}\quad  In \cref{tab: finetue_gym}, our SSL with skeleton input outperforms most of the fully supervised methods and achieves the best performance among unsupervised pre-train methods with RGB input. For the same input modality, our performance is lower than the fully supervised method PYSKL\cite{duan2022revisiting} (with the skeleton as input) by about 1.4\%, because the PYSKL adopts stacks of visual heatmaps of skeleton joints as input while we only use human skeleton coordinates. This validates the promising discriminative ability of our skeleton sequence representation. 

\noindent\textbf{Results on Diving48 Dataset.}\quad  Our SSL with skeleton input outperforms some fully supervised methods. Although our SSL is not pre-trained on additional large-scale pre-training action datasets in \cref{tab: diving48}, it still achieves competitive performance among unsupervised pre-train methods. This validates that our pre-training model SkeletonMAE can learn discriminative skeleton sequence representation. 

The results on FineGym and Diving48 validate that our SkeletonMAE has a promising ability to enhance the feature representation of skeleton sequence by comprehensively perceiving the underlying topology of actions, and the SSL can learn discriminative action representation. 

\noindent\textbf{Results on NTU 60 and NTU 120 Datasets.}\quad In \cref{tab: finetune_ntu}, for NTU 60 X-sub and NTU 60 X-view, compared with unsupervised pre-train methods, our SSL outperforms the current state-of-the-art method Colorization \cite{yang2021skeleton} by 4.8\% and 1.6\%, respectively. Our SSL is also competitive compared with fully supervised methods, outperforming the first six fully supervised methods on NTU 60 X-sub. For NTU 120 X-sub and NTU 120 X-set, our SSL outperforms the previous best-unsupervised pre-train method 3s-PSTL \cite{zhou2023self}  by 3.5\% and 3.1\%, respectively. Our SSL is superior compared with some fully supervised methods on NTU 120 X-sub and NTU 120 X-set. These results show that our SSL can learn discriminative skeleton representation from large-scale action recognition datasets due to the promising generalization ability of our SkeletonMAE.

\subsection{Ablation Studies} 
\label{sec: abl}
In this section, we analyze the contributions of essential components and hyper-parameters of our model. Note that unless otherwise specified, all experiments are verified on the FineGym dataset with masking body sub-region 3.
 
\noindent\textbf{Whether to load pre-trained model or not.}\quad  To explore the effectiveness of loading the pre-trained SkeletonMAE encoder, we find that the accuracy is 86.3 without loading the pre-trained SkeletonMAE encoder (randomly initialized weights). As \cref{table:abla2}(a) shows, loading the pre-trained model is always better than not loading it. This validates that our SkeletonMAE can learn more comprehensive and generalized representations for unlabeled fine-grained actions by reconstructing the skeleton joint features.

\noindent\textbf{GIN layers in SkeletonMAE.}\quad \cref{table:abla}(a) shows the performance of using different GIN layers in the SkeletonMAE encoder. The performance is the best when $L_{D}=3$.

\begin{table}[!t]
    \centering
	\begin{minipage}{0.4\linewidth}
		\centering
        \footnotesize
		\setlength{\tabcolsep}{1.0mm}{
  \vspace{-10pt}
\begin{tabular}{c|c}
\hline
$L_{D}$    & Mean Acc.     \\ \hline
1          & 89.6          \\
2          & 90.7          \\
\textbf{3} & \textbf{91.2} \\
4          & 90.9          \\ \hline
\end{tabular}
    }
    \centerline{(a)}
	\end{minipage}\begin{minipage}{0.6\linewidth}
		\centering
        \footnotesize
		\setlength{\tabcolsep}{1.0mm}
{        \vspace{-4pt}
       \begin{tabular}{l|c}
\hline
Method                & Mean Acc.     \\ \hline
GraphCL \cite{you2020graph}               & 86.5          \\
JOAO \cite{you2021graph}                 & 88.7          \\
\textbf{Ours(SkeletonMAE)} & \textbf{91.2} \\ \hline
\end{tabular}}
        \centerline{(b)}
	\end{minipage}
 
\vspace{6pt}
\begin{minipage}{1.0\linewidth}
		\centering
        \small
		\setlength{\tabcolsep}{1.0mm}
{        \vspace{-1pt}
\begin{tabular}{l|cccccc}
\hline
\# Masked Body Part & 0    & 1    & 2    & 3    & 4    & 5    \\ \hline
 GAT \cite{velickovic2017graph}             & 86.8 & 88.1 & 88.9 & 89.5 & 89.4 & 90.0 \\
 GCN \cite{kipf2016semi}             & 87.6 & 88.9 & 89.3 & 90.6 & 89.5 & 90.5 \\
GIN \cite{xu2018powerful}                & \textbf{88.6} & \textbf{89.5} & \textbf{90.2} & \textbf{91.2} & \textbf{90.3} & \textbf{91.2} \\ \hline
\end{tabular}}
        \centerline{(c)}
	\end{minipage}
 \vspace{-10pt}
        \caption{(a) Mean accuracy of using the different number of GIN layers in SkeletonMAE encoder. (b) Comparison results with the contrastive learning method as the pre-trained encoder.  (c) The results of using different backbones in SkeletonMAE under each masked body part are compared.}
        \vspace{-15pt}
        \label{table:abla}
\end{table}

\noindent\textbf{Comparison with contrastive learning methods.}\quad  
To verify the superior ability of our SkeletonMAE when conducting skeleton sequence pre-training, we compare our SkeletonMAE with different contrastive learning methods GraphCL and JOAO. As shown in \cref{table:abla}(b), our SkeletonMAE achieves the best performance. Besides, we visually compare the action representations of SkeletonMAE and GraphCL by PCA, as shown in \cref{fig: visu}(a) and \cref{fig: visu}(b). Compared to GraphCL, the skeleton representation of our SkeletonMAE appears to have a larger inter-class variance and smaller intra-class variance. This validates that our SkeletonMAE can comprehensively capture the human pose and obtain discriminative skeleton sequence representation. We observe similar patterns in all other classes but visualize only five categories for simplicity.

\noindent\textbf{Backbones and masked body parts in SkeletonMAE.}\quad As shown in \cref{table:abla}(c),  we show the accuracy of our SSL with different SkeletonMAE backbones and different masked body parts in SkeletonMAE. It can be seen that GIN is always better than both GAT and GCN under the same masked body part. This is because that GIN provides a better inductive bias for graph-level applications. Thus, it is more suitable for learning more generalized skeleton representations. Additionally, we can see that masking body sub-regions 3 and 5 are both optimal across all backbones, which demonstrates the importance of reconstruction of human limbs in action recognition. 
\begin{table}[!t]
    \centering
	\begin{minipage}{1.0\linewidth}
		\centering
		\setlength{\tabcolsep}{0.96mm}{
\begin{tabular}{l|cccc}
\hline
\textit{M}        & 1    & 2    & 3    & 4    \\ \hline
Mean Acc.    & 89.1(\textcolor[RGB]{147,72,51}{$\uparrow$}\textcolor[RGB]{147,72,51}{2.8}) & 90.6(\textcolor[RGB]{147,72,51}{$\uparrow$}\textcolor[RGB]{147,72,51}{4.3}) & \textbf{91.2(\textcolor[RGB]{147,72,51}{$\uparrow$}\textcolor[RGB]{147,72,51}{4.9})} & 91.0(\textcolor[RGB]{147,72,51}{$\uparrow$}\textcolor[RGB]{147,72,51}{4.7}) \\ \hline
\end{tabular}
    }
    \centerline{(a)}
	\end{minipage}
\vspace{6pt}
\begin{minipage}{1.0\linewidth}
		\centering
        \small
		\setlength{\tabcolsep}{0.1mm}
{        \vspace{2pt}
\begin{tabular}{c|c|cccc}
\hline
\multicolumn{2}{l|}{\# Masked Joints Number}                         & 5                                                   & 9                                                         & 12                                                                 & 15                                                                 \\ \hline
\multicolumn{2}{l|}{Ratio of Mask Joints}                   & 30\%                                                & 50\%                                                      & 70\%                                                               & 90\%                                                               \\ \cline{3-6} 
    \multicolumn{2}{l|}{Accuracy of SSL} & 89.7                                                & 90.3                                                      & 89.9                                                               &90.1                                             \\ \hline
\multirow{2}{*}{\thead{\\  Masked Body Part }} 
& \textbf{High}         & \begin{tabular}[c]{@{}c@{}}91.8\\ ($\mathcal{V}_3$,$\mathcal{V}_5$)\end{tabular} & \begin{tabular}[c]{@{}c@{}}91.2\\ ($\mathcal{V}_0$,$\mathcal{V}_3$,$\mathcal{V}_5$)\end{tabular} & \begin{tabular}[c]{@{}c@{}}91.0\\ ($\mathcal{V}_1$,$\mathcal{V}_2$,$\mathcal{V}_3$,\\ $\mathcal{V}_4$,$\mathcal{V}_5$)\end{tabular} & \begin{tabular}[c]{@{}c@{}}90.8\\ ($\mathcal{V}_0$,$\mathcal{V}_1$,$\mathcal{V}_2$,\\ $\mathcal{V}_3$,$\mathcal{V}_5$)\end{tabular} \\ 
\cline{2-6} 
& \textbf{Low}          & \begin{tabular}[c]{@{}c@{}}91.1\\ ($\mathcal{V}_2$,$\mathcal{V}_4$)\end{tabular} & \begin{tabular}[c]{@{}c@{}}90.1\\ ($\mathcal{V}_0$,$\mathcal{V}_1$)\end{tabular}    & \begin{tabular}[c]{@{}c@{}}91.0\\ ($\mathcal{V}_1$,$\mathcal{V}_2$,$\mathcal{V}_3$,\\ $\mathcal{V}_4$,$\mathcal{V}_5$)\end{tabular} & \begin{tabular}[c]{@{}c@{}}90.2\\ ($\mathcal{V}_0$,$\mathcal{V}_1$,$\mathcal{V}_3$,\\ $\mathcal{V}_4$,$\mathcal{V}_5$)\end{tabular} \\ 
\hline
\end{tabular}}
        \centerline{(b)}
	\end{minipage}
 \vspace{-15pt}
        \caption{(a) Results of four SSL variants. \textcolor[RGB]{147,72,51}{$\uparrow$} represents the accuracy improvement relative to the random initialization of SkeletonMAE in SSL.  (b) The comparison of our body part based masked and the random masked strategies. $\mathcal{V}_0$-$\mathcal{V}_5$: \textcolor[RGB]{121,121,121}{Head},  \textcolor[RGB]{253,222,187}{Torso},  \textcolor[RGB]{249,64,63}{Left arm}, \textcolor[RGB]{255,160,62}{Right arm},  \textcolor[RGB]{183,86,215}{Left leg}, \textcolor[RGB]{2,142,85}{Right leg}.  }
        \vspace{-17pt}
        \label{table:abla2}
\end{table}


\noindent\textbf{Variants of SSL.}\quad 
To evaluate whether our pre-trained SkeletonMAE is effective across different skeleton action recognition models, we set the different number of STRL layers ($M=1,2,3,4$) to obtain four variants of the SSL. As shown in \cref{table:abla2}(a), all SSL variants outperform the random initialization
of SkeletonMAE in SSL, which validates our body part masking strategy indeed improves the discriminative ability of skeleton feature by learning action-sensitive visual concepts. Additionally, three-layer STRL is the best due to the good compromise between efficiency and computational cost. Moreover, it also validates that our SkeletonMAE generalizes well across different models.

\noindent\textbf{Skeleton Masked Strategy.}\quad 
In \cref{table:abla2}(b), our masked body part strategy is fairly compared with the random masked strategy under the same masked joint conditions. Our method is better than the random mask method across all settings. As mentioned in \cref{sec:Pre-tr}, our masking strategy is action-sensitive and reconstructs specific limbs or body parts that dominates the given action class and is suitable for real-world skeleton-based action recognition.

\noindent\textbf{Transferability of the SkeletonMAE across datasets.}\quad 
As shown in \cref{fig5:trans}(a), we pre-train SkeletonMAE on the FineGym dataset and then fine-tune it on NTU 60 X-sub and NTU 120 X-sub datasets.
Compared with the method that uses the same dataset for pre-training and fine-tuning, our SkeletonMAE achieves better performance across all masked strategies when conducting dataset transfer. This shows that the SkeletonMAE can learn generalized skeleton representation and effectively transfer the strong representation ability to other datasets.

\subsection{Visualization Analysis}
\begin{figure}[t]
  \centering
   \vspace{-20pt}
  \begin{center}
    \includegraphics[scale=0.45]{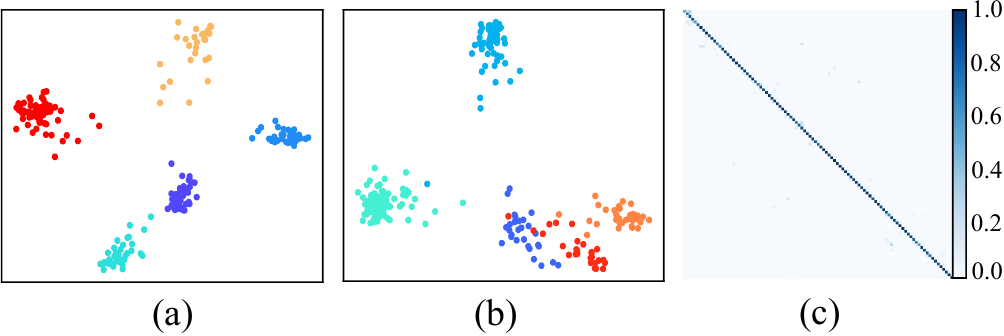}
  \end{center}
  \vspace{-8pt}
  \hfil  \vspace{-22pt}\caption{ (a) and (b) are 2d-PCA of SkeletonMAE and GraphCL as pre-trained encoder representations. We randomly select five action classes for 2d-PCA visualization, (c) Confusion matrix for fine-grained action recognition.}
\vspace{-6pt}
  \label{fig: visu}
\end{figure}

\begin{figure}
\begin{minipage}{0.48\linewidth}
  \centerline{\includegraphics[width=4.2cm]{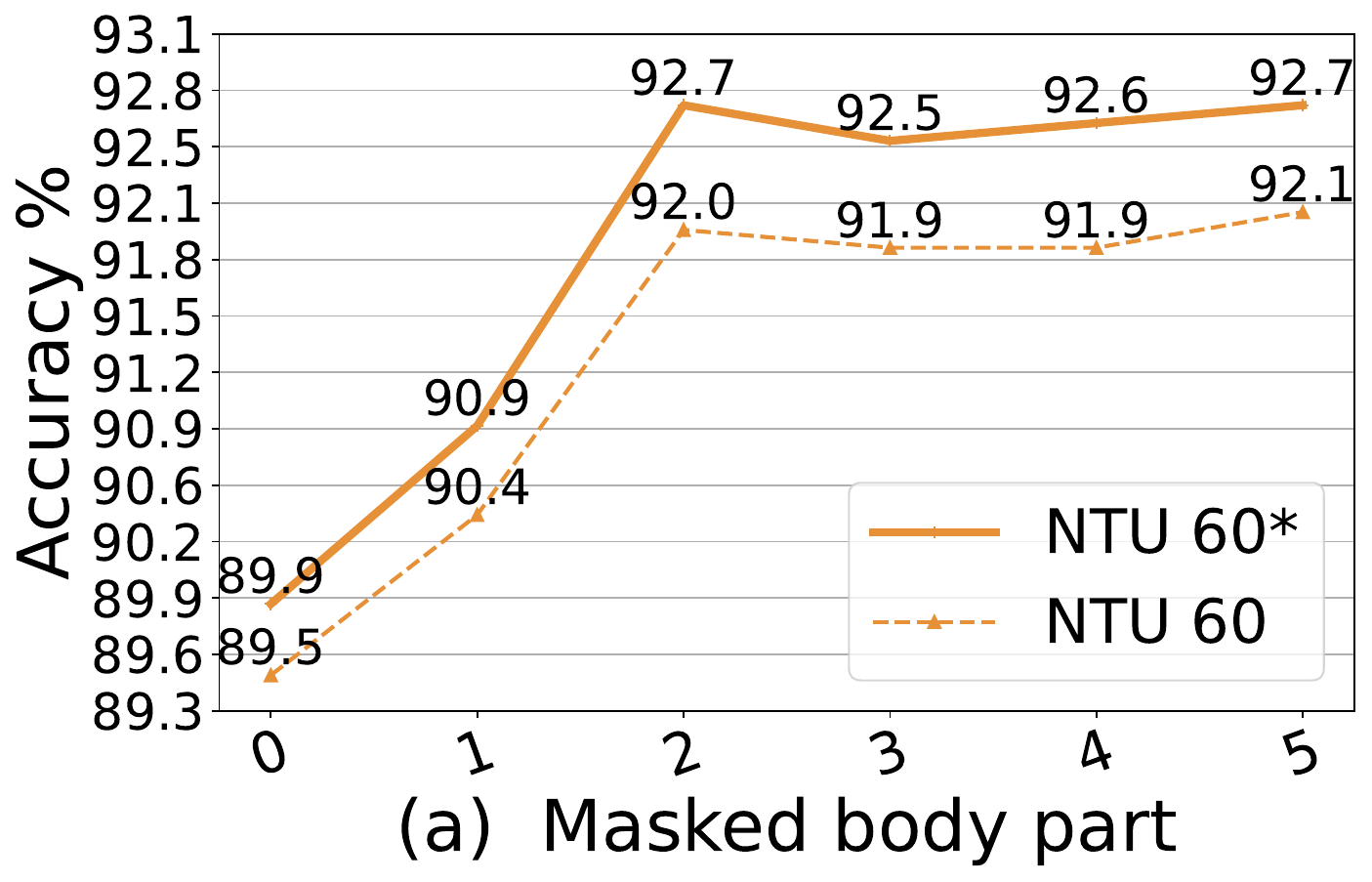}}
\end{minipage}
\hfill\begin{minipage}{0.48\linewidth}
  \centerline{\includegraphics[width=4cm]{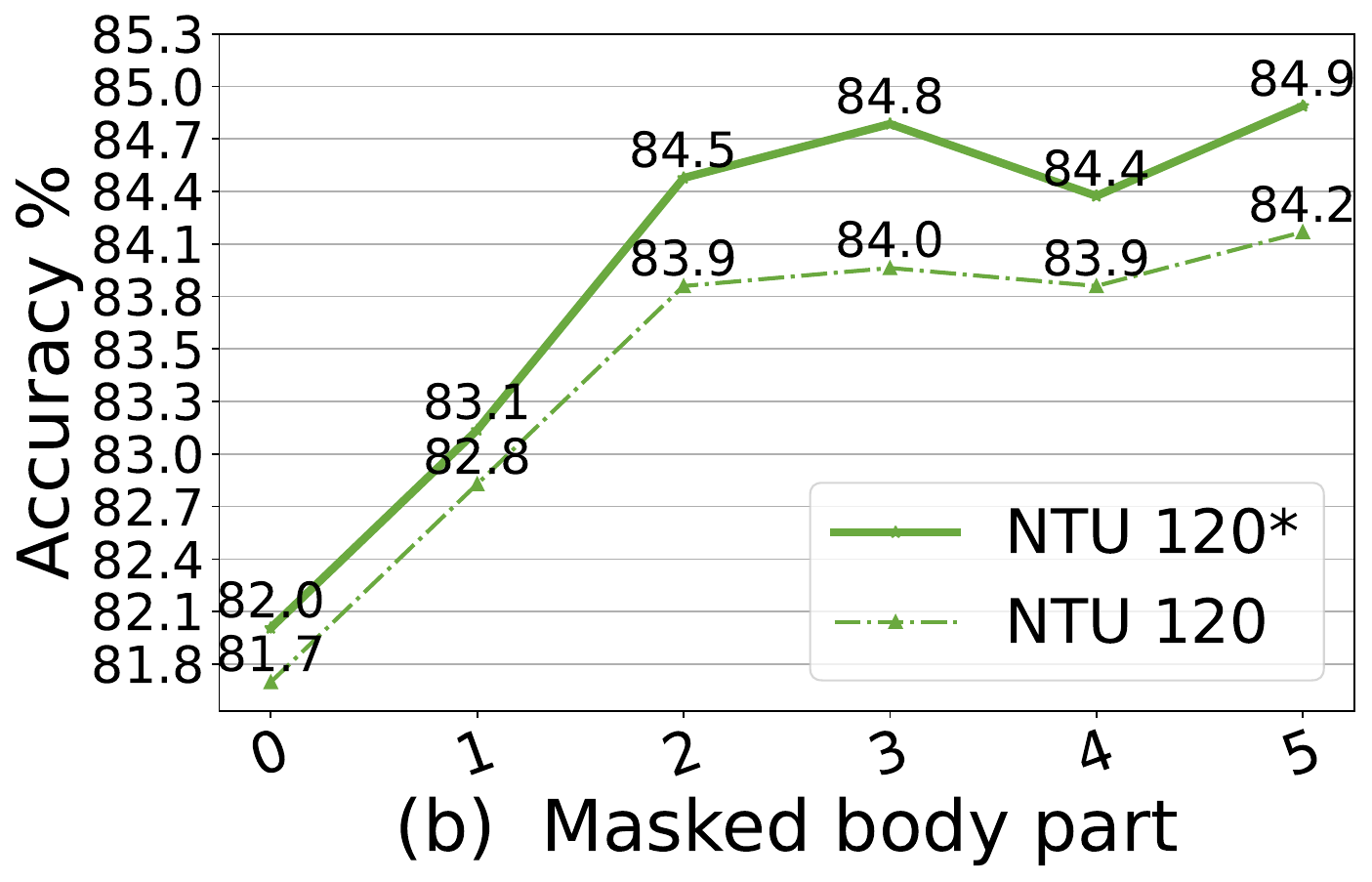}}
\end{minipage}
 \vspace{-10pt}
\caption{(a) The accuracies with mask body part of 0-5 on the NTU 60 X-sub dataset, and (b) the accuracies on the NTU 120 X-sub dataset. `\textsuperscript{*}' means SkeletonMAE encoder pre-trained on the FineGym dataset.
}
 \vspace{-6pt}
\label{fig5:trans}
\end{figure}

\begin{figure}[!t]
  \centering
  \begin{center}
    \includegraphics[scale=0.35]{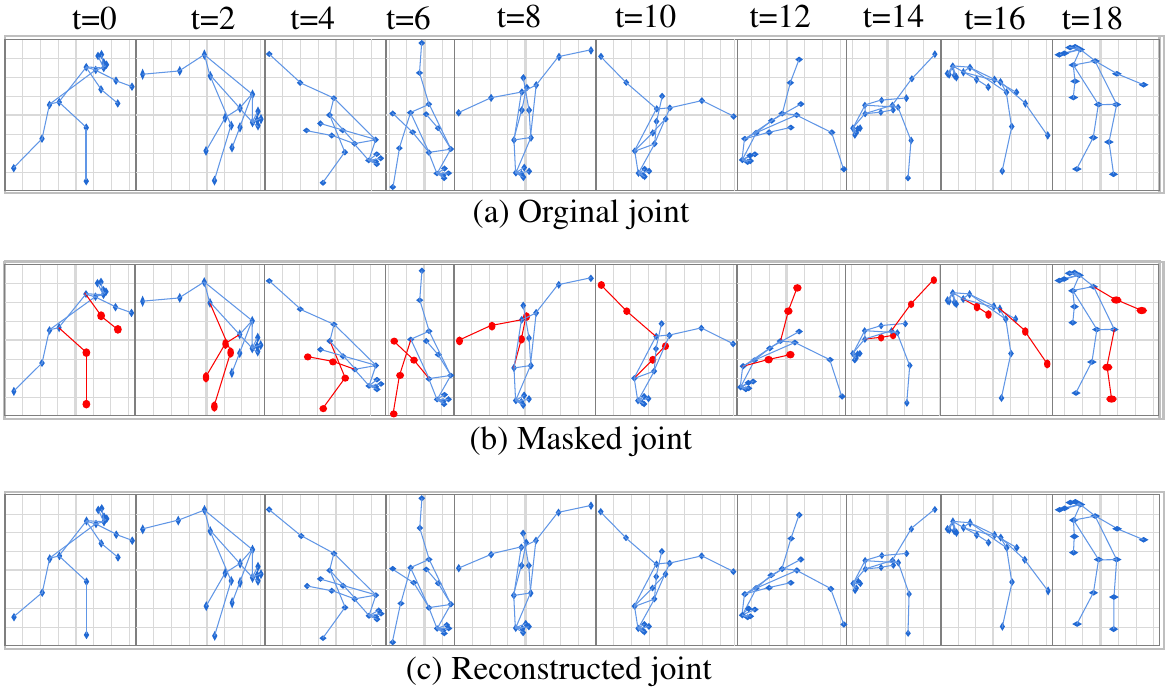}
  \end{center}
  \hfil  \vspace{-30pt}\caption{Visualization results for skeleton sequence of ``aerial walkover forward" action on FineGym dataset. (a) The input skeleton sequence. (b) Masked skeleton sequence (masked parts are 3 and 5). (c) Reconstructed skeleton sequence.}
\vspace{-15pt}
  \label{fig:masked}
\end{figure}

\cref{fig:masked} shows the reconstruction process of the skeleton sequence by SkeletonMAE. From the same frame, the difference between the reconstructed skeleton sequence and the original skeleton sequence is slight, but overall the human body structure is reserved. This shows that the SkeletonMAE has good spatial representation learning ability. \cref{fig: visu}(c) shows that our SSL works well for fine-grained action recognition tasks on the FineGym dataset. More visualization results are in Supplementary Section C. 

\vspace{-5pt}
\section{Conclusion}
\vspace{-5pt}
\label{sec:conclusion}
In this paper, we propose an efficient skeleton sequence learning framework, SSL, to learn discriminative skeleton-based action representation. To comprehensively capture the human pose and obtain skeleton sequence representation, we propose a graph-based encoder-decoder pre-training architecture, SkeletonMAE, that embeds skeleton joint sequence into GCN and utilize the prior human topology knowledge to guide the reconstruction of the underlying masked joints and topology. Extensive experimental results show that our SSL achieves SOTA performance on four benchmark skeleton-based action recognition datasets.

{\small
\bibliographystyle{ieee_fullname}
\bibliography{egbib}
}

\end{document}